\documentclass{article}
\usepackage{graphicx}
\usepackage{physics}
\usepackage{amsmath}
\usepackage{amssymb}
\usepackage{amsthm}
\usepackage{mathtools}
\usepackage{a4wide}
\usepackage{authblk}
\usepackage{xurl}
\usepackage{tikz} 
\usepackage{pgfplots}
\usepackage{hyperref}
\pgfplotsset{compat=1.18}
\usepackage{multirow}

\usepackage[letterpaper,top=2cm,bottom=2cm,left=3cm,right=3cm,marginparwidth=1.75cm]{geometry}
\usepackage{subcaption}
\usepackage{float}
\usepackage{totcount}
\usepackage{breakcites}
\usetikzlibrary{trees}
\usepackage[normalem]{ulem}

\title{A Unified Perspective on Optimization in Machine Learning and Neuroscience: From Gradient Descent to Neural Adaptation}

\author[]{Jesús García Fernández}
\author[]{Nasir Ahmad}
\author[]{Marcel van Gerven}

\date{}

\affil[1]{Department of Machine Learning and Neural Computing, Donders Institute for Brain, Cognition and Behaviour, Radboud University, Nijmegen, the Netherlands}

\begin{document}

\maketitle

\begin{abstract}

Iterative optimization is central to modern artificial intelligence (AI) and provides a crucial framework for understanding adaptive systems.
This review provides a unified perspective on this subject, bridging classic theory with neural network training and biological learning.
Although gradient-based methods, powered by the efficient but biologically implausible backpropagation (BP), dominate machine learning, their computational demands can hinder scalability in high-dimensional settings.
In contrast, derivative-free or zeroth-order (ZO) optimization feature computationally lighter approaches that rely only on function evaluations and randomness.
While generally less sample efficient, recent breakthroughs demonstrate that modern ZO methods can effectively approximate gradients and achieve performance competitive with BP in neural network models.
This ZO paradigm is also particularly relevant for biology.
Its core principles of random exploration (probing) and feedback-guided adaptation (reinforcing) parallel key mechanisms of biological learning, offering a mathematically principled perspective on how the brain learns.
In this review, we begin by categorizing optimization approaches based on the order of derivative information they utilize, ranging from first-, second-, and higher-order gradient-based to ZO methods.
We then explore how these methods are adapted to the unique challenges of neural network training and the resulting learning dynamics.
Finally, we build upon these insights to view biological learning through an optimization lens, arguing that a ZO paradigm leverages the brain's intrinsic noise as a computational resource.
This framework not only illuminates our understanding of natural intelligence but also holds vast implications for neuromorphic hardware, helping us design fast and energy-efficient AI systems that exploit intrinsic hardware noise.

\end{abstract}

\section{Introduction}
\label{sec:introduction}

Finding the most effective configuration of a complex system hinges on optimization, that is, the minimization of an objective function~\cite{nocedal1999numerical}.
In machine learning, where even small improvements in this process can yield significant benefits in a given task, efficient and reliable optimization algorithms are crucial~\cite{sra2011optimization}.
Specifically, when training neural networks (NNs), this involves navigating vast parameter spaces to drive the system towards a desired goal by minimizing a loss function~\cite{Goodfellow2016}.
This process is frequently framed as an unconstrained optimization problem, where the goal is to find the optimal set of parameters without explicit boundaries or constraints~\cite{vazquez2025optimization}.
However, the inherent complexities of these models, including high dimensionality, non-convex loss landscapes, and the large volume of data required for model fitting, render this task challenging.

Formally, unconstrained optimization can be stated as the minimization of a scalar-valued objective function
\begin{align*}
    \theta^*  = \min _{\theta} f(\theta)
\end{align*}
where $\theta \in \mathbb{R}^n$ represents a vector of $n$ parameters (e.g., weights and biases in a neural network) and $f\colon \mathbb{R}^n \rightarrow \mathbb{R}$ assigns a performance or cost value to each configuration of these parameters. Iterative optimization algorithms typically seek to refine the parameter vector $\theta$ at each time step through an update rule 
\begin{align*}
\theta \gets  \theta - \eta d^*
\end{align*}
where $d^* \in \mathbb{R}^n$ is the chosen descent direction, designed to reduce the value of $f(\theta)$, and $\eta$ is the step size, which determines the magnitude of the update.

A principled way to determine the optimal descent direction is to minimize a local approximation of $f$ around $\theta$. This approximation, denoted $m_k(d) \approx f(\theta + d)$, is typically derived from a $k$-th order Taylor expansion of $f$ around $\theta$. To ensure the step $d$ remains in a region where this approximation is valid, the model $m_k(d)$ often incorporates a penalty on the step size, or the minimization is constrained to a local neighborhood (a ``trust region''). The optimal descent direction is thus given by
\begin{align}
d^* = \min_{d} m_k(d)
\label{eq:taylormodel}
\end{align}
where $k$ indicates the order of the approximation and the model $m_k(d)$ is understood to be regularized. For a first-order model ($k=1$), this constrained minimization yields the direction of steepest descent, making $d^*$ parallel to the gradient $\nabla f(\theta)$. For $k>0$, the approximation employs gradient information, and we refer to the resulting methods as gradient-based optimization methods. These methods are described in more detail in Section~\ref{sec:gradient_based_opt}.

\begin{figure}
    \centering
    \includegraphics[width=\linewidth]{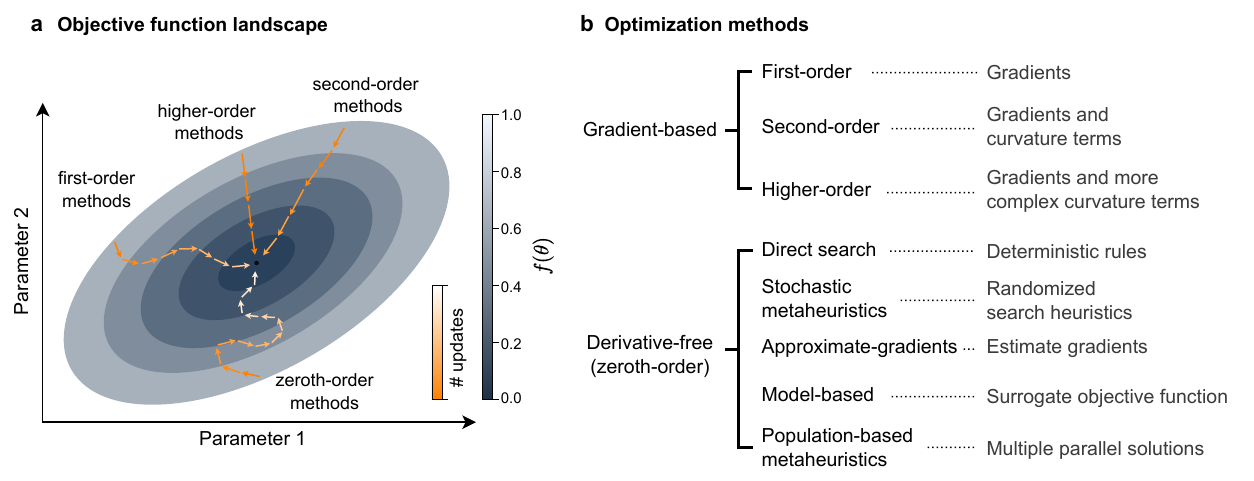}
    \caption{\textbf{A visual guide to optimization methods.} \textbf{(a)} Objective function landscape depicting example trajectories for different categories of methods. The higher the order, the more information it considers, and the fewer updates it needs to find the minimum. \textbf{(b)} Taxonomy tree categorizing the broad groups of optimization methods discussed in the review.}
    \label{fig:Fig1}
\end{figure}

However, not all optimization contexts readily permit the computation of gradients. This has led to the development of derivative-free or zeroth-order (ZO) optimization methods~\cite{conn2009introduction, rios_derivative-free_2013, larson_derivative-free_2019, liu_primer_2020}, which rely solely on objective function evaluations. These methods are particularly valuable when dealing with non-differentiable components, optimizing black-box systems, or when gradient computations are too costly. However, this flexibility comes at the cost of slower convergence and higher sample complexity, especially in high‑dimensional settings. In Figure \ref{fig:Fig1}a, a visual diagram shows how optimization approaches of different orders navigate the landscape of the objective function. The evaluation‑based updates of ZO methods also align naturally with biologically plausible learning rules. In contrast, the biological implementation of backpropagation (BP) is considered highly implausible~\cite{Crick1989}; a challenge that has inspired the search for alternative learning paradigms~\cite{lillicrap2020backpropagation}. The brain's reliance on local activity and global feedback signals provides a natural substrate for ZO-like optimization, a possibility that we explore in detail in Section \ref{bio-learning}. 

Understanding the spectrum of optimization approaches, ranging from derivative-dependent to derivative-free methods, as shown in Figure~\ref{fig:Fig1}b, is crucial. In a landscape fragmented by specialized methodologies, a holistic perspective can reveal deeper connections. This review offers a unique perspective, examining their strengths, limitations, and conceptual overlaps, as well as their implications for biological learning. Viewing biological learning through an optimization lens can offer profound insights into the brain's adaptive capabilities and extend our understanding of both natural and artificial intelligence.

The remainder of this paper is organized as follows. Section 2 explores gradient-based optimization methods, which leverage first- and higher-order derivatives of the objective function. Section 3 then shifts focus to derivative-free optimization techniques, which operate without direct access to gradient information. In Section 4, we contextualize these optimization paradigms specifically within the domain of NNs. Building on this foundation, Section 5 then argues that biological learning may be better understood as a form of zeroth-order optimization, a perspective that provides insights into its adaptive capabilities. Finally, Section 6 summarizes our findings and explores the implications of this unifying framework for both natural and artificial intelligence.

\section{Gradient-based optimization}
\label{sec:gradient_based_opt}

This section examines optimization methods that leverage explicit derivative information from the objective function to guide parameter updates. By measuring how the objective function changes with respect to the parameters, these methods identify promising descent directions. 
Consider again Eq.~\eqref{eq:taylormodel}. 
The Taylor expansion can be expressed as \begin{equation}
m_k(d)=\sum_{i=0}^k \frac{1}{i!} \nabla^i f\left(\theta\right)\left[d^{\otimes i} \right] + g(d)
\label{eq:taylor}
\end{equation}
where $\nabla^i f\left(\theta\right)$ denotes the $i$-th derivative tensor of $f$ at $\theta$, $\nabla^i f\left(\theta\right)\left[d^{\otimes i}\right]$ denotes plugging in $d$ in all $i$ slots of the tensor (multilinear form)~\cite{nesterov2018lectures}, and $g(d)$ denotes a regularization. Gradient-based optimization encompasses a spectrum of techniques that span first-order schemes, second-order methods, and higher-order methods, depending on the choice of $k$. Below, we explore each of these categories, detailing their core mechanisms, advantages, and limitations.

\subsection{First-order methods}

First-order optimization methods rely exclusively on the gradient (the first derivative) of the objective function to determine the descent direction. Conceptually, these approaches correspond to minimizing the first-order Taylor approximation ($k=1$), guiding the search along the direction of steepest descent. Minimizing $m_1(d)=f(\theta)+\nabla f(\theta)^T d + d^T d$ yields the updates defined below. 

\textbf{Gradient descent (GD)}~\cite{nocedal1999numerical} is the most fundamental first-order optimization algorithm. It iteratively updates the parameters by moving in the direction opposite to the gradient of the objective function, as this indicates the direction of steepest local descent. The optimal descent direction is given by
\begin{align*}
    d^* = \nabla f(\theta) .
\end{align*}
This method directly implements the principle of minimizing the first-order Taylor approximation $f(\theta+d)$, where $d=\nabla f(\theta)$. While simple and intuitive, gradient descent requires computing the gradient over the entire dataset, which can be prohibitively expensive for large‑scale problems. Moreover, although Euclidean gradient steps guarantee a monotonic decrease in the objective under smoothness assumptions, they often “zig‑zag” through narrow valleys of the loss landscape or even stall at saddle points, resulting in slow progress even with exact gradient information.

\textbf{Stochastic gradient descent (SGD)}~\cite{robbins_stochastic_1951, boyd_convex_2004} is developed as a more scalable alternative to gradient descent. SGD addresses the high computational cost of full‑batch gradient descent by replacing the exact gradient $\nabla f(\theta)$ with an unbiased estimate 
computed on a small random subset (minibatch) of $\mathcal{B}$ samples. The optimal descent direction is approximated by
\begin{equation}
    d^* \approx \frac{1}{\left|\mathcal{B}\right|} \sum_{i \in \mathcal{B}} \nabla_\theta \mathcal{L}\left(x_i ; \theta\right)
\end{equation}
where $\mathcal{L}\left(x_i ; \theta\right)$ denotes the loss on sample $x_i$. By trading off some variance in $g$ for dramatically lower per‐step cost, SGD makes it feasible to train large‐scale models, with per update complexity $O(n)$ in the parameter dimension $n$.

While effective, SGD can converge slowly in ravines (areas where the surface is steeper in one dimension than another) or stall at saddle points. To address these limitations, several enhancements have been developed. Momentum methods~\cite{polyak_methods_1964, nesterov_method_1983} can accelerate SGD in relevant directions by adding a fraction of the update vector from the previous time step to the current update. Gradient scaling and normalization (e.g.,~\cite{tieleman_lecture_2012}) can prevent exploding gradients and stabilize training. Modern methods, like Adam~\cite{kingma_adam_2017}, Adagrad~\cite{duchi_adaptive_2011}, NAdam~\cite{dozat_incorporating_2016} and AdamW~\cite{loshchilov2017decoupled}, often combine these techniques.

\subsection{Second-order methods}
\label{sec:second_order_methods}

Second-order optimization methods advance beyond first-order approaches by incorporating information about the curvature of the objective function's landscape. This is achieved by utilizing second-order derivatives, such as the Hessian matrix or the Fisher information matrix. As a result, these methods take more direct and efficient steps towards the optimum, conceptually corresponding to minimizing the second-order Taylor approximation ($k=2$). Minimizing the approximation $m_2(d)=f(\theta)+\nabla f(\theta)^T d+\frac{1}{2} d^T \nabla^2 f(\theta) d$ yields Newton-type updates defined below.

\textbf{Newton's method}~\cite{ypma_historical_1995} is a classic second-order optimization algorithm that uses the Hessian matrix $\nabla^2 f(\theta)$ to account for local curvature and compute a more direct update than gradient descent. At each iteration, it inverts the Hessian and applies it to the update, giving the descent direction
\begin{align*}
    d^* = [\nabla^2 f(\theta)]^{-1} \nabla f(\theta).
\end{align*}
Its primary advantage is its rapid, quadratic convergence when sufficiently close to a local minimum. However, its $O(n^2)$ memory requirement for storing the $n\times n$ Hessian and $O(n^3)$ cost of matrix inversion make it impractical for high‑dimensional NNs. Moreover, if the Hessian is not positive definite, which is common in non‑convex settings, the update can point toward saddle points or even increase the objective.  

\textbf{Quasi-Newton's methods}~\cite{nocedal1999numerical} emerged as a practical alternative to mitigate the high computational burden in Newton's method. They sidestep the cost of forming and inverting the Hessian by building an approximate (inverse) Hessian from successive gradient evaluations. By updating this curvature estimate at each step, they retain many advantages of second-order information while using only first-order data.

Prominent examples include the Broyden–Fletcher–Goldfarb–Shann algorithm (BFGS)~\cite{broyden_convergence_1970, fletcher_new_1970, goldfarb_family_1970, shanno_conditioning_1970}, and its limited‐memory variant L-BFGS~\cite{liu1989limited}. These methods often attain superlinear convergence in practice, with each iteration costing in the order of $O(n)$, but can struggle when the objective is nonconvex or noisy, as curvature estimates may become unreliable.

\textbf{Natural gradient}~\cite{amari_natural_1998, zhang_fast_2019, pascanu_revisiting_2014} takes a different approach by introducing a geometry-aware second-order approach. Instead of using the Hessian matrix, which assumes a Euclidean structure, this method utilizes the Fisher information matrix $I(\theta)$ to define a metric of the parameter space. This is a positive semi-definite matrix and therefore avoids some issues of second-order methods. This matrix captures the intrinsic geometry of the statistical model defined by the parameters, allowing one to take steps that are invariant to reparameterizations of the model. The descent direction is given by
\begin{align*}
    d^* = I(\theta)^{-1} \nabla f(\theta).
\end{align*}
Natural gradient methods often exhibit faster convergence and improved stability in ill-conditioned or highly curved loss landscapes, making them particularly useful in neural network training, where the parameter space has a non-Euclidean geometry. However, computing and inverting the Fisher matrix remains computationally expensive in high-dimensional settings, with a typical cost of $O(n^3)$, which limits their scalability in practice.

\textbf{Modern curvature approximations} have been developed to make second-order optimization more practical for NNs. While second-order methods offer strong theoretical benefits, their direct application is hindered by the high computational and memory costs of computing and inverting the full Hessian or Fisher Information Matrix. To overcome this, recent work has focused on efficient approximation techniques that capture useful curvature information without incurring prohibitive costs.

These methods often exploit the layered structure of NNs to simplify the curvature matrices or compute only the necessary matrix–vector products for use in iterative solvers. Notable examples include Kronecker-Factored Approximate Curvature (K-FAC)~\cite{martens_optimizing_2015}, which approximates the Fisher matrix using a block-diagonal Kronecker-factored structure, enabling efficient natural gradient updates. EK-FAC~\cite{george2018fast} builds on this by incorporating empirical Fisher estimates and low-rank corrections to improve approximation quality. Shampoo~\cite{gupta_shampoo_2018} applies preconditioning using approximate matrix square roots to incorporate second-order information at scale. A recent extension, SOAP \cite{vyas_soap_2025}, combines Shampoo with the adaptive Adam optimizer to improve stability and performance. Another example, projected natural gradient descent (PRONG)~\cite{desjardins_natural_2015}, improves efficiency by decorrelating neural activations to approximate natural gradient updates more effectively. In a similar vein, decorrelated backpropagation (DBP)~\cite{dalm2024efficient, ahmad2024correlations} proposes adding a learnable decorrelation process to the inputs of each layer, aligning the gradient more closely with the natural gradient and leading to faster convergence.

\subsection{Higher-order methods}

Higher-order optimization methods~\cite{nesterov2021implementable, cartis2024efficient} extend the principles of gradient-based optimization by utilizing derivative information beyond the second order (i.e., third-order derivatives and above). These methods capture the geometry of the objective function's landscape more accurately, theoretically enabling even more precise steps towards an optimum. For example, for $k=3$, minimizing the approximation $$m_3(d)= f(\theta)+\nabla f(\theta)^T d+\frac{1}{2} d^T \nabla^2 f(\theta) d+\frac{1}{6} \nabla^3 f(\theta)[d, d, d]$$ leads to higher-order tensor methods~\cite{nesterov2021implementable}.

Their primary advantage lies in accelerated convergence rates, surpassing the quadratic convergence of Newton's method. For example, cubic regularization algorithms~\cite{nesterov2006cubic}, a well-known class of third-order methods, can achieve cubic convergence rates under certain conditions. This makes them highly attractive for problems where extreme precision or few iterations are critical.

However, the computational and memory costs associated with higher-order derivatives quickly become prohibitive as the order increases. Computing and storing higher-order tensors requires astronomical resources for high-dimensional problems such as neural network training. As a result, these methods find limited practical application, remaining mostly theoretical or confined to small-scale applications.

\section{Derivative-free (zeroth-order) optimization}
\label{sec:derivative_free_opt}

This section introduces a distinct class of optimization algorithms known as derivative-free or ZO optimization methods, which operate without access to the gradient of the objective function. Unlike gradient-based approaches, ZO methods rely solely on evaluating the objective function $f(\theta)$ at different points in the parameter space to infer promising descent directions. This paradigm is particularly advantageous in scenarios where derivative information is unavailable or impractical to compute. Despite their versatility, ZO methods generally incur higher sample complexity (by requiring more function evaluations to achieve a similar level of convergence) and exhibit slower convergence, particularly in high-dimensional spaces.

The following subsections will provide an overview of the diverse landscape of ZO optimization techniques. We categorize the methods presented here by their underlying strategies. While some methods explore the parameter space with deterministic, systematic rules, others introduce stochasticity to explore the landscape more broadly, or even employ parallel multi-point evaluation strategies. This section also highlights how some of these methods can be formally linked to gradient-based principles. These links are particularly relevant, as some of these principles can also be connected to biological learning, a subject we will explore further in later sections.

\subsection{Direct search methods}

These methods are characterized by their systematic exploration of the parameter space, using a set of deterministic rules to select new points for evaluation~\cite{kochenderfer2019algorithms}. They operate solely on function values without building a surrogate model or approximating the gradient.

\textbf{Coordinate search}~\cite{luo1992convergence}, also known as coordinate descent, iteratively minimizes the objective function along one coordinate direction at a time while keeping all other parameters constant. This process continues by cycling through all dimensions until a stopping criterion is met. Although simple to implement, its performance can be poor in high-dimensional problems, especially when the parameters are strongly correlated.

\textbf{Hooke-Jeeves}~\cite{hooke1961direct} alternates between an exploratory phase, where a series of searches are performed at a specific step size around the current point along each dimension, and a pattern-moving phase, where a step is taken in the direction that yields the largest improvement. If no improvement is found at the search points, the step size is reduced. This process repeats until the step size is sufficiently small, making it an effective method for small-dimensional problems.

\textbf{Mesh adaptive direct search (MADS)}~\cite{audet2006mesh}, also known as generalized pattern search, alternates between exploring a set of search directions from the current point and pattern-moving towards an improving point, similar to Hooke-Jeeves. However, MADS can search in arbitrary directions, rather than only in coordinate directions. If no improvement is found during the exploration, the direction scale is reduced. This method is also effective in small-dimensional problems.

\textbf{Nelder--Mead (simplex)}~\cite{nelder_simplex_1965} operates by maintaining a geometric figure called a simplex, defined by $n+1$ vertices (in an $n$-dimensional space), each corresponding to an evaluation point. At each iteration, the vertex with the highest function value is replaced with a new candidate that yields a lower function value via geometric transformations (reflection, expansion, contraction, and shrinkage). This process allows the simplex to elongate along flat regions and contract near minima. However, it scales poorly to high-dimensional problems.

\textbf{Powell’s method}~\cite{powell1964efficient} operates by performing a series of one-dimensional line searches. The method maintains a set of conjugate search directions, which are updated at each iteration. It performs a line search along each direction to find the minimum, and then constructs a new direction from the difference between the starting and ending points of the line searches. This method requires only function evaluations and line searches, but can be costly in problems with high dimensions.

\textbf{Divided rectangles (DIRECT)}~\cite{jones1993lipschitzian} operates by iteratively dividing the search space into smaller hyper-rectangles ($n$-dimensional generalization of a rectangle) based on the objective function values, trying to narrow down the search space to the global optimum. This method balances global search by exploring new areas and local search by refining existing promising areas, using a heuristic inspired by potential Lipschitz constants. Its complexity grows exponentially with dimension, making it suitable mainly for low-dimensional problems.

\subsection{Stochastic metaheuristics}

Unlike deterministic direct search methods, these approaches use randomness to explore the search space. They operate solely on function values too, and are particularly effective at escaping local optima.

\textbf{Random search}~\cite{rastrigin1963convergence} is the most basic form of stochastic optimization, where the evaluation points are randomly sampled from the search space. It is guaranteed to find the global optimum given an infinite number of samples. However, it is extremely sample-inefficient and has no memory of past evaluations. Although very simple to implement, it is ineffective in high-dimensional problems.

\textbf{Simulated annealing}~\cite{kirkpatrick1983optimization} is a method inspired by the metallurgical process of annealing, where a metal is heated and then slowly cooled, making it more workable. Here, the temperature is used to control the degree of stochasticity during the process. The process starts with a high temperature, allowing for a wide exploration of the parameter space. It randomly samples new evaluation points and accepts solutions that reduce the objective function with a certain probability. The `temperature' is then slowly decreased, reducing the probability of accepting new solutions and yielding local improvements. Its practical performance is highly dependent on the cooling schedule and the proposal distribution (used to find new evaluation points), making it unsuitable for high-dimensional problems.

\subsection{Approximate-gradient methods}

These methods explicitly aim to approximate the gradient of the objective function without requiring its analytical form. By doing so, they can leverage the efficiency and convergence properties of gradient-based optimization algorithms (to a certain extent) when the true gradient is unavailable or too expensive to compute.

\textbf{Finite-difference approximations}~\cite{kiefer_stochastic_1952} are the most direct approach to approximating the gradient. The core idea is to estimate the partial derivative of the function with respect to each parameter by evaluating the function at a slightly perturbed point. Given the parameter vector $\theta$ and the objective function $f$, the partial derivative with respect to a specific parameter can be approximated in several ways. The forward difference approximates the partial derivative with respect to the $i$-th parameter as
\begin{align*}
    \frac{\partial f}{\partial \theta^{(i)}} \approx \frac{f\left(\theta+h e_i\right)-f(\theta)}{h}
\end{align*}
where $e_i$ is the standard basis vector for the $i$-th dimension and $h>0$ is a small scalar step size. A more accurate approximation is the central difference, which uses evaluations on both sides of the parameter as
\begin{align*}
    \frac{\partial f}{\partial \theta^{(i)}} \approx \frac{f\left(\theta+h e_i\right)-f(\theta-h e_i)}{2h}.
\end{align*}

Despite their simplicity, computing the full gradient requires $n$ and $2n$ function evaluations per step (where $n$ is the number of parameters), respectively, which makes them computationally prohibitive in high-dimensional problems. A notable extension of this idea is the Kiefer–Wolfowitz algorithm~\cite{kiefer1952stochastic}, specifically designed for noisy or stochastic objective functions. It employs a perturbation size that decreases over time, which allows it to find a minimum even in the presence of noise.

These methods serve as a bridge between purely derivative-free and gradient-based techniques by constructing approximate gradients usable in standard descent algorithms. This has led to the development of several ZO counterparts to popular optimizers, such as ZO stochastic gradient descent (ZO-SGD)~\cite{ghadimi_stochastic_2013}, ZO sign-based stochastic gradient descent (ZO-signSGD)~\cite{liu_signsgd_2019}, and ZO stochastic variance reduced gradient (ZO-SVRG)~\cite{liu_zeroth-order_2018, ji_improved_2019}. These algorithms mimic the behavior of their counterparts without using explicit gradients. For a more comprehensive exploration of these advanced methods, readers can refer to~\cite{liu_primer_2020}.

\textbf{Simultaneous perturbation stochastic approximation (SPSA)}~\cite{spall_multivariate_1992, spall2005introduction} is a more efficient method that estimates a full gradient vector with only two function evaluations, regardless of the number of parameters. It mirrors the central finite difference by applying perturbations in both positive and negative directions. However, SPSA simultaneously perturbs all parameters with a single random vector $\xi \sim p$, obtaining an estimation of the full gradient as
\begin{align*}
    \nabla f(\theta) \approx \frac{f(\theta+h \xi) - f(\theta -h \xi) }{2} (h\xi)^{-1}
\end{align*}
where each component $\xi_i$ is sampled independently from a distribution $p$ (commonly a Rademacher distribution, i.e. $\xi_i = \pm 1$), and $h>0$ is a small step size. The element-wise inverse $\xi^{-1}$ corrects for the random sign and magnitude of each coordinate, ensuring an unbiased estimate of the true gradient. 

Although SPSA is far more scalable than finite-difference methods, its implicit stochasticity yields slower convergence compared to deterministic gradient-based methods. As the dimensionality scales up, the variance of its estimate increases, limiting its applicability to very high-dimensional problems.

\subsection{Model-based methods}

These methods build and maintain a surrogate model of the objective function, which is cheaper to evaluate than the actual function. The optimization algorithm then utilizes this model to select the next point to sample from the real function, finding the optimum with the fewest evaluations possible.

\textbf{Bayesian optimization (BO)}~\cite{garnett2023bayesian} is a sample-efficient optimization strategy designed for expensive-to-evaluate black-box functions, which leverages information from past evaluation points. It operates by iteratively involving two components. The first component is a probabilistic surrogate model of the objective function $f$, which is typically a Gaussian process. The model is updated using Bayes' theorem~\cite{bayes1763lii}, such that given a set of $t$ observations $D_t = \{(\theta_1, f(\theta_1)), \ldots, (\theta_t, f(\theta_t))\}$, the Gaussian process provides a posterior distribution $P(f \mid D_t)$, from which the mean $\mu(\theta)$ and the uncertainty (variance) $\sigma^2(\theta)$ can be derived. The second component is an acquisition function that uses this information to determine the next point to sample. This balances the need to explore areas with high uncertainty (exploration) and exploit areas that are likely to contain the optimum (exploitation).

By focusing evaluations where they are most informative, this method often finds the global optimum in fewer steps than other similar methods. This makes it particularly well-suited for hyperparameter tuning in machine learning~\cite{snoek2012practical, shahriari2015taking}, where each function evaluation (i.e., validating a neural network) is computationally costly. Although being extremely sample-efficient, Bayesian optimization scales very poorly to high-dimensional problems. Extensions of this method, such as the use of embeddings~\cite{wang2016bayesian}, high-dimensionality variants~\cite{snoek2015scalable, papenmeier2025bayesian}, and multi-fidelity variants~\cite{kandasamy2017multi}, partially mitigate these scalability issues.

\textbf{Radial basis function (RBF) networks}~\cite{buhmann2000radial}, in contrast to the former method, act as a deterministic surrogate model of the objective function $f$. Iteratively, new points are chosen by explicitly optimizing this surrogate model, the true function is evaluated at those points, and the data is then used to refine the surrogate model. 

The values of a RBF depend only on the distance from a central point (a point previously evaluated), and a separate RBF is placed at each evaluated point. The surrogate model, $\hat f$, is then constructed as a weighted sum of these individual basis functions as
\begin{align*}
    \hat f(\theta) = \sum_{t=1}^T \lambda_s \phi(\| \theta - \theta_t \|)
\end{align*}
where $\theta_t$ is a previously evaluated point at time $t$, $T$ is the total number of previous points, $\phi$ is the RBF kernel (e.g., Gaussian, multi-quadric), and $\lambda_s$ are the learned coefficients for each RBF.

These methods are highly effective in low-dimensional problems where the surrogate model can be constructed accurately from a limited number of samples. The main challenge is that the number of RBFs grows with the number of evaluation points, which can make the model difficult to manage in large-scale problems.

\textbf{Trust-region methods}~\cite{levenberg1944method, yuan2015recent} build a surrogate model $\hat f$ of the objective function around the current point $\theta$ and optimize this model within a neighborhood (the trust region with radius $\delta$). The next evaluation point $\theta^\prime$ is then obtained by solving
\begin{align*}
    & \min _{\theta} \hat f(\theta) 
\end{align*}
subject to $\| \theta - \theta^\prime\| \le \delta$. The radius of the trust region $\delta$ is adapted according to how well the model predicts the actual objective. For that, a ratio $\alpha$ is calculated, comparing the actual improvement $f(\theta) - f(\theta^\prime)$ with the predicted improvement $f(\theta) - \hat f(\theta^\prime)$, such that
\begin{align*}
    \alpha = \frac{f(\theta) - f(\theta^\prime)}{f(\theta) - \hat f(\theta^\prime)} .
\end{align*}
If $\alpha$ is close to 1, the model is a good approximation, and $\delta$ is increased for the next iteration to allow for a larger step. If $\alpha$ is small or negative, the model is not accurate. In this case, $\delta$ is shrunk to ensure the next step is taken in a region where the approximation is more likely to be valid.

Unlike line search methods, which fix a search direction and then find a step size, trust-region methods determine both the direction and length of the step simultaneously, which can improve their convergence time. However, these methods scale poorly with dimensionality due to the computational cost of the exact subproblem, making them inefficient in high-dimensional problems.

\subsection{Population-based metaheuristics}

These methods operate on a set of candidate solutions (i.e., evaluation points) simultaneously. Unlike previous single-point methods, which iteratively improve a single solution, these approaches leverage the collective intelligence of a population. This parallel exploration of the search space gives them an advantage in scenarios where single-point methods might get trapped in local optima.

\textbf{Evolutionary algorithms (EA)}~\cite{back1993overview, yu2010introduction, vikhar2016evolutionary} are inspired by the principles of biological evolution and natural selection. They maintain a population of solutions and iteratively improve them through operators that mimic natural processes such as selection, recombination, and mutation. The main algorithms are described below.

\textit{Genetic algorithms} (GA)~\cite{mitchell1998introduction, katoch2021review} maintain a population of candidate solutions, often encoded as a bit-string representation (chromosomes). In each generation, solutions are evaluated based on their fitness (best objective function $f$ values). Fitter solutions are more likely to be selected to serve as ``parents'' for the next generation. New solutions are created through crossover, by combining genetic material from two parents, and mutation, by randomly altering parts of a solution (e.g., replacing bits in the chromosomes). 

\textit{Genetic programming} (GP)~\cite{koza1994genetic, o2009riccardo} is a specialized form of a GA in which the solutions are computer programs or functions. Instead of optimizing a vector of parameters, GP evolves the structure of a program itself, typically represented as a tree. The evolutionary operators, crossover and mutation, are designed to manipulate these program trees.

\textit{Evolution strategies} (ES)~\cite{beyer_evolution_2002, slowik_evolutionary_2020} focus more on mutation and selection processes than crossover. The primary mechanism involves generating a population of $N$ candidate solutions, where each candidate $\theta_i$ is created by adding random noise to the current point, such that $\theta_i = \theta + \sigma \epsilon_i$, where $\epsilon_i \sim \mathcal{N}(0, I)$ is a random perturbation vector. The parameter vector $\theta$ is then updated with a weighted average of the perturbations, with weights determined by their respective fitness value. Remarkably, this update has been shown to converge to finite-difference approximations of the gradient, such that
\begin{align*}
    \nabla f(\theta) \approx\frac{1}{N \sigma^2} \sum_{i=1}^N f\left(\theta_i\right) \epsilon_i
\end{align*}
indicating that ES approximate gradient descent~\cite{zhang_relationship_2017, lehman_es_2018, raisbeck_evolution_2019}. 

\textit{Differential evolution} (DE)~\cite{storn1997differential, das2010differential} is a simpler method that uses population-based mutation. For each solution $\theta_i$, a new candidate solution is created by combining three randomly chosen solutions ($\theta_1, \theta_2, \theta_3$), such that $\theta_{new} = \theta_1 + \beta(\theta_2 - \theta_3)$, where $\beta$ is a scaling factor. This new solution is then crossed over with the original solution, and the better of the two survives to the next generation.

\textbf{Swarm intelligence algorithms}~\cite{eberhart2001swarm, kennedy2006swarm} are inspired by the collective behavior of decentralized, self-organized systems in nature. They rely on the interactions between individual agents and their environment. The main algorithms are described below.

\textit{Particle swarm optimization} (PSO)~\cite{kennedy1995particle, poli2007particle} is a method where each candidate solution is a particle moving through the search space. The movement of each particle is guided by its own best-known position $p^*$ and the best-known position of the entire swarm $g^*$. Each particle $i$ moves through the search space, with its velocity $v_i$ and position $\theta_i$ updated in each iteration as
\begin{align*}
    v_i & \leftarrow \omega v_i + c_1 r_1 (p^*_{i} - \theta_i) + c_2 r_2 (g^* - \theta_i) \\
    \theta_i & \leftarrow \theta_i + v_i
\end{align*}
where $\omega, c_1, c_2$ are coefficients and $r_1, r_2 $ are random numbers drawn from the uniform distribution $\mathcal{U}(0, 1)$. This approach mimics the use of momentum to accelerate convergence towards minimums.

\textit{Ant colony optimization} (ACO)~\cite{dorigo1996ant, dorigo2007ant} is a method inspired by the foraging behavior of ants to find optimal paths. Here, artificial ants stochastically navigate a graph or network, leaving artificial pheromone trails on the edges they cross. The paths used more frequently by successful ants accumulate more pheromones, which in turn increases the probability of more ants following those paths. The pheromone update rule for a path segment (from $i$ to $j$) is typically given by $\tau_{i,j} = (1-\rho)\tau_{i,j} + \sum_k \Delta \tau_{i,j}^k $, where $\rho$ is the rate of evaporation and $\Delta \tau_{i,j}^k$ is the amount of pheromone deposited by ant $k$. This positive feedback loop allows the algorithm to find the shortest or most efficient path over time.

Other nature-inspired algorithms~\cite{yang2010nature} have been proposed after the success of algorithms like PSO and ACO. They follow the same general principle of agents exploring a search space based on local interactions and collective information. For example, the \textit{Firefly algorithm}~\cite{yang2009firefly} mimics the flashing behavior of fireflies. It models a population of agents (represented by fireflies) that move towards brighter (better) solutions. The light intensity of a firefly represents the objective function value. \textit{Cuckoo search}~\cite{yang2009cuckoo} is inspired by the brood parasitism of cuckoos. This method generates new solutions (represented by eggs) by using a random walk (Lévy flights) and replacing inferior solutions with better ones (like the cuckoo's strategy of laying eggs). \textit{Bee colony optimization}~\cite{karaboga2005idea} is inspired by the foraging behavior of honey bees. It divides the population into three groups of bees (scout, onlooker, and employed bees) to balance exploration of new areas and exploitation of known promising areas. \textit{Bat algorithm}~\cite{Yang2010Bat} mimics the echolocation of bats. Each bat is a potential solution that moves through the search space searching for prey using a combination of a fixed-frequency pulse and a variable pulse rate to find and exploit optimal solutions.

\textbf{Hybrid population metaheuristics}~\cite{blum2003metaheuristics, boussaid2013survey} combine population-based search with complementary local or heuristic strategies. Their main goal is to achieve a better balance between diversification (exploring the search space to locate promising regions) and intensification (performing focused local searches to refine solutions within those regions). We cover some of the main algorithms below.

\textit{Memetic algorithms} (MA)~\cite{moscato1989evolution, neri2012memetic} combine an evolutionary framework with one or more local search components. The term ``meme'', in cultural evolution, represents a piece of individual knowledge that is improved during its lifetime. In optimization, it is a strategy that an individual solution can undergo to improve its fitness during iterations. These methods can be classified into two styles of lifetime learning~\cite{ku1997exploring}: Lamarckian learning, where lifetime improvements are directly stored in the genome and inherited by offspring, and Baldwinian learning, where lifetime improvements enhance an individual’s fitness and chances of being selected, but are not encoded in the genome or passed on genetically. These methods can directly link the population-based search to gradient approximation, as the local search can be a gradient-based method.

Other stochastic heuristics combine a population-based framework with other stochastic search strategies, maintaining a collection of solutions and using stochastic processes to guide their evolution. This hybridization allows for a more sophisticated control over the exploration-exploitation trade-off. For example, \textit{estimation of distribution algorithms} (EDAs)~\cite{larranaga2001estimation}, instead of using genetic operators like crossover and mutation, build a probabilistic model of promising solutions in the current population and sample from this model to generate new solutions. \textit{Scatter search and path relinking algorithms} (SS\&PR)~\cite{glover1999scatter} work on a small, high-quality ``reference set'' of solutions and generate new solutions by creating paths between them. This is a highly structured form of intensification. \textit{Cultural algorithms (CA)}~\cite{reynolds1994introduction}, inspired by human cultural evolution, use a dual-level system with a ``population space'' for individual solutions and a ``belief space'' that stores the collective knowledge of the population. The belief space guides the population's search, providing a high-level, knowledge-driven system for managing the search process.

\section{Optimization in neural networks}

Optimization lies at the heart of neural network training, adjusting the network parameters to drive the system toward a desired goal~\cite{bishop1995neural, Goodfellow2016}. Although we have introduced a wide range of general optimization methods in previous sections, applying them to neural networks presents a unique set of challenges that distinguish this task from traditional optimization.

Unlike classical optimization problems, neural network training is inherently high-dimensional. It involves minimizing an objective function defined over massive datasets, rather than a fixed objective function, and over a parameter space containing millions to billions of parameters, which results in a complex, non-convex loss landscape. These unique characteristics have led to the development of specialized techniques for efficient gradient computation, as well as a range of adapted optimization methods that are robust to this environment. This section will delve into these unique aspects, bridging the gap between general optimization theory and practical application in NNs.

In addition, the unexpected success in optimizing these overparameterized models, which have more parameters than data points and are organized in terms of complex computational graphs, has given rise to surprising phenomena like double descent~\cite{belkin2019reconciling} and grokking~\cite{power2022grokking}. We will further explore these intriguing learning dynamics at the end of this section.

\subsection{Neural network objective function and gradient computation}

At its core, neural network training is an optimization problem aimed at finding parameters $\theta$ (weights and biases) that minimize a specific objective function. Unlike the objective function $f(\theta)$ considered in previous sections, the objective function of a neural network, termed the loss function, depends on the input data. Formally, for a single data instance $(x_i, y_i)$, where $x_i$ is an input sample and $y_i$ is the corresponding target, the loss is expressed as $l(\theta;x_i, y_i)$. When considering a set of data $\mathcal{D} = \{(x_i, y_i)\}^N_{i=1}$ of $N$ samples, the overall loss of this set $L(\theta, \mathcal{D})$ is typically computed as
\begin{align*}
    L(\theta, \mathcal{D}) = \frac{1}{N} \sum_{i=1}^{N} l(\theta; x_i, y_i) .
\end{align*}
In practice, computing the full sum of losses is not feasible due to the large size of the datasets. Instead, the loss and its gradient are calculated using mini-batches, small and randomly sampled data subsets. We will elaborate on this in the following section. To minimize this loss function, the standard strategy is to use gradient-based methods to compute the derivatives of the loss function with respect to the parameters and update those parameters accordingly. The unique characteristics of NNs have led to the development of specialized techniques for efficient gradient computation, which are detailed below.

\textbf{Automatic differentiation (AD)}~\cite{baydin2018automatic} is the default method for computing gradients in NNs. AD is a powerful computational technique that accurately evaluates the derivatives of functions defined by computer programs. AD computes gradients by iteratively applying the chain rule to the elementary operations represented within the network's computational graph. Unlike symbolic differentiation~\cite{griewank2008evaluating}, it does not produce unwieldy expressions and, unlike numerical differentiation~\cite{press2007numerical}, it provides exact derivatives (up to machine precision). This makes AD the fundamental engine for training NNs, enabling efficient gradient information acquisition. There are two main AD modes. \textit{Reverse mode AD} performs a forward pass to compute and store all intermediate values, followed by an additional backward pass that propagates the derivative of the final output back through the graph towards the input nodes. This computes the full gradient at a fixed cost, regardless of the input dimension, making it highly efficient when the number of outputs is small. \textit{Forward mode AD} computes and propagates forward the derivatives of each intermediate value with respect to the inputs, avoiding the need for a backward pass. In each operation, the local derivative is combined with the accumulated sensitivities of previous steps. To obtain the full gradient, forward mode AD requires as many passes as input dimensions, making it more efficient when the number of inputs is small. 

\begin{figure}
    \centering
    \includegraphics[width=\linewidth]{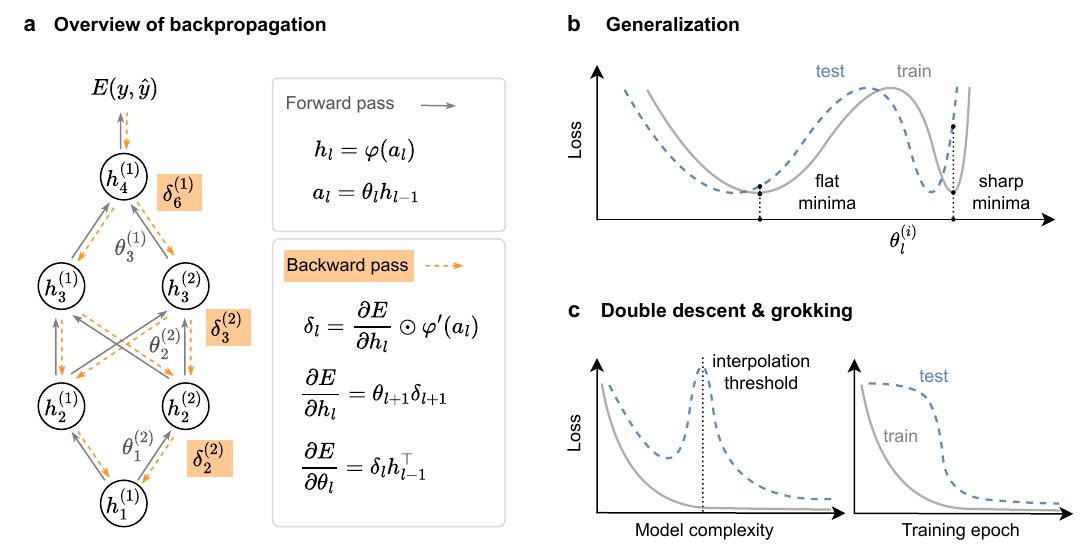}
    \caption{\textbf{Learning dynamics in neural networks.} \textbf{(a)} Overview of the computational graph and operations of backpropagation in an $L$-layer neural network. $E$ represents the output error between the network output $\hat y$ and a given target, $\delta_l^{(i)}$ is the node-specific error of layer $l$ and node $i$, $\theta_{l}^{(i)}$ represents a single parameter in the network in layer $l$ connecting to node $i$, $\varphi$ is an activation function, $h_i$ represent the nodes outputs and $a_i$ represent the pre-activations. Only some of the values of $\theta$ and $\delta$ are shown for simplification. \textbf{(b)} Loss landscapes and how different minima affect generalization. \textbf{(c)} Double descent and grokking phenomena.}
    \label{fig:Fig2}
\end{figure}

\textbf{Backpropagation}~\cite{linnainmaa1970representation, werbos1974beyond, rumelhart_learning_1986} is a practical implementation of reverse mode AD, where the forward pass builds the computational graph of tensor operations and the backward pass traverses it in reverse to accumulate gradients. It leverages the network's layered structure to apply the chain rule backward efficiently. Figure \ref{fig:Fig2}a provides an overview of its computational graph and the operations carried out in each phase. BP’s cost scales roughly linearly with the forward pass operations, independent of parameter count, making it the workhorse of modern neural network training. It is essential to note that BP itself only computes these exact gradients, which are then utilized by optimization algorithms (e.g., SGD, Adam, etc.) to update the network's parameters. Modern frameworks, such as PyTorch~\cite{paszke2019pytorch}, JAX~\cite{jax2018github}, and TensorFlow~\cite{tensorflow2015-whitepaper}, automate graph construction and both passes, eliminating manual derivative work. For recurrent neural networks (RNNs), however, a specialized form of BP is required. Backpropagation through time (BPTT)~\cite{werbos1990backpropagation} applies the same principles of reverse-mode AD to the unfolded computational graph of an RNN. It treats the network at each time step as a distinct layer, allowing the gradient to be computed over the entire sequence. However, this can be computationally expensive and can suffer from vanishing or exploding gradients over long sequences. An alternative to BPTT is real-time recurrent learning (RTRL) \cite{williams1989learning}, which computes gradients forward in time at each step, enabling online learning without the need for a full sequence pass. Nevertheless, RTRL has a higher computational complexity per step, making it less practical for large NNs.

\textbf{Forward gradients}~\cite{baydin_gradients_2022} is a practical implementation of forward mode AD, which propagates random perturbations forward. It computes directional derivatives along these random directions and utilizes them to construct an unbiased estimate of the full gradient. Each directional derivative is obtained in a single forward pass and is then aggregated to form the gradient estimate, avoiding the need for a backward step. Similar to BP, forward gradients only computes gradients, which are then utilized by optimization algorithms to update the network’s parameters. However, computing the exact gradient requires as many passes as the input dimension. Instead, often an (inexact) gradient estimate is measured and used for training. This often makes the method impractical for training NNs, which often have high-dimensional inputs. Although extensions such as activity perturbed methods~\cite{ren_scaling_2022}, low variance techniques~\cite{bacho_low-variance_2024}, auxiliary network biasing~\cite{fournier_can_2023}, and multi-point estimates~\cite{flugel_beyond_2024} have been proposed to alleviate these costs, forward gradient approaches remain uncommon in standard neural network training.

\subsection{Applications of optimization methods in neural networks}

Having introduced the main optimization methods (Sections \ref{sec:gradient_based_opt} and \ref{sec:derivative_free_opt}), we now examine their applications within NNs. First, we consider how gradient-based methods, the standard approach to training NNs, are adapted to address specific challenges in this context. Then, we will explore derivative-free techniques as a less common but increasingly relevant alternative, highlighting their advantages and showcasing their recent successes in training large NNs from scratch.

\subsubsection{Gradient-based methods in neural networks}

First‑order gradient methods dominate neural network training due to their scalability to billions of parameters, computational efficiency, and robustness to the noisy, high‑dimensional loss landscapes of deep NNs. They use exact gradients from BP (or, less commonly, estimates from forward gradients) to iteratively update parameters. Although Section \ref{sec:gradient_based_opt} covered their mechanics, applying these methods in practice requires key adaptations. One of the most important is mini-batching, which computes gradients on small data subsets to provide a noisy but efficient estimate of the full gradient. Given the massive size of the datasets, mini-batching is a necessary technique that strikes a balance between computational efficiency and regularization. It provides a sufficient level of noise in the gradient updates, which helps the model escape sharp local minima and leads to better generalization. Furthermore, mini-batching is designed to leverage the parallel processing power of modern hardware like GPUs. Learning rate schedules~\cite{darken1992learning} dynamically adjust step sizes to navigate complex landscapes, while adaptive optimizers (e.g., Adam~\cite{kingma_adam_2017}, Adagrad~\cite{duchi_adaptive_2011}) assign per-parameter rates based on historical gradients, which is particularly useful for sparse data or non‑uniform loss landscapes.

Despite their robustness, first‑order methods must still contend with loss landscapes full of flat regions, saddle points, and local minima; though the stochastic updates of mini-batching help escape saddle points. The initial configuration of the network parameters also plays a crucial role in the success of the optimization process. Poor initialization can lead to issues like as vanishing or exploding gradients, stalling learning, or causing unstable updates. Initialization schemes, such as Xavier/Glorot~\cite{glorot2010understanding} and Kaiming/He~\cite{he2015delving}, are designed to preserve activation and gradient scales across layers. Regularization techniques, such as L2 weight decay~\cite{van2017l2}, dropout~\cite{srivastava2014dropout}, and batch normalization~\cite{ioffe2015batch}, reshape the loss landscape to improve generalization and indirectly ease optimization. Architectural innovations like skip connections~\cite{he2016deep} further smooth the landscape~\cite{li2018visualizing}, while methods like sharpness-aware minimization~\cite{foret2020sharpness} prioritize finding flat minima to improve generalization.

In contrast, second-order and higher-order methods are generally prohibitive for large NNs due to their enormous computational and memory requirements. Constructing, storing, and inverting Hessian matrices or higher-order tensors becomes infeasible as their size scales quadratically with parameter count. However, their principles, particularly the use of curvature information, have driven practical approximations, such as K-FAC~\cite{martens_optimizing_2015}, EK-FAC~\cite{george2018fast}, Shampoo~\cite{gupta_shampoo_2018}, PRONG~\cite{desjardins_natural_2015}, or DBP~\cite{dalm2024efficient}, which are covered at the end of Section \ref{sec:second_order_methods}.

\subsubsection{Zeroth-order methods in neural networks}

While gradient-based methods remain the dominant approach for neural network training, their limitations in specific contexts have sparked growing interest in ZO optimization methods. These limitations include computational and memory challenges as networks scale to massive sizes (despite the efficiency of BP, modern architectures are pushing boundaries), incompatibility with non-differentiable components such as spiking neurons~\cite{tavanaei2019deep}, unsuitability for black-box optimization where internal states are inaccessible~\cite{rios_derivative-free_2013}, and constraints posed by neuromorphic hardware~\cite{schuman2017survey}. Moreover, some ZO methods offer an alternative perspective that aligns more naturally with the principles of biological plausibility~\cite{lillicrap2020backpropagation}, relying on local perturbations paired with a global feedback signal, a topic we will revisit in later sections.

The diverse landscape of ZO optimization (introduced in Section \ref{sec:derivative_free_opt}) has found applications in NNs, but its effectiveness is limited when scaling to their high-dimensional parameter spaces. This limitation arises from the high variance of the gradient estimates, a core mechanism in many ZO methods. This variance leads to slow convergence in high dimensions, making many ZO methods impractical for training NNs from scratch. Despite these challenges, various ZO categories have been applied to NNs. Methods in the categories \textit{direct search}, \textit{non-gradient stochastic metaheuristics}, and \textit{model-based methods} are primarily used for hyperparameter optimization~\cite{snoek2012practical, bergstra2012random, liashchynskyi2019grid}, where the search space is much lower-dimensional.

Methods in the category \textit{approximate-gradient methods} are particularly relevant and effective here, as they directly mimic the core mechanism of gradient descent. Weight perturbation (WP)~\cite{jabri_weight_1991, cauwenberghs_fast_1992, werfel_learning_2003} is a popular practical implementation of these methods, used to train NNs, in which a small amount of random noise is added to the parameters via a Gaussian vector $\epsilon \sim \mathcal{N}(\vec{0}, I)$. Then, the change in the loss function is measured to approximate the gradient as
\begin{align*}
    \nabla f(\theta) \approx \frac{f(\theta+ \sigma^2 \epsilon) - f(\theta) }{\sigma^2} \epsilon
\end{align*}
where $\sigma^2 > 0$ denotes the variance of the distribution. Unlike SPSA, WP does not invert the perturbation vector (in this case $\epsilon$). This results in an estimation of the gradient of a Gaussian‑smoothed version of the objective function $f$, which reduces variance and improves robustness to noise. WP shares the same scalability benefits and drawbacks as SPSA. A single function evaluation per step (plus baseline), regardless of dimensions, makes it far more scalable than finite difference methods. However, its accuracy also degrades in very high dimensions, given the variance in its gradient estimate. Note that although WP was proposed differently and without a strong theoretical link to gradient descent, it is now understood to be mathematically equivalent to the forward gradient method described above.

Node perturbation (NP)~\cite{flower_summed_1992, werfel_learning_2003} is a specialized extension of WP that reduces the gradient estimation variance by perturbing neuronal pre-activations instead of individual weights. By significantly reducing the effective dimensionality of the perturbation space, NP produces more efficient gradient estimates. A recent extension~\cite{dalm_effective_2023} further refines this by considering the propagated effects of perturbations through the network, using differences in neural activity. However, even these advancements struggle to completely mitigate the problem of growing variance in high-dimensional spaces, which ultimately limits their use for training large NNs~\cite{hiratani_stability_2022, werfel_learning_2003, zuge_weight_2023}.

Nevertheless, recent breakthroughs demonstrated the potential of ZO methods to train large NNs in supervised learning settings, achieving performance that is competitive with traditional gradient-based approaches. For example, DeepZero~\cite{chen_deepzero_2023} applies coordinate-wise finite differences to estimate gradients for each parameter, leveraging massive parallelization across parameter blocks. It achieves competitive performance with BP when training ResNet architectures~\cite{he2016deep} on CIFAR-10~\cite{krizhevsky2009learning}, scaling to models with up to 300 million parameters. However, the need to perturb and evaluate each parameter independently, despite parallelization, can significantly increase computational cost, potentially reducing the efficiency advantages typically associated with ZO methods. Extending this idea to recurrent architectures, Chaubard and Kochenderfer~\cite{chaubard2025scaling} apply a similar finite-difference strategy with highly parallel computation to train RNNs with up to a billion parameters. Their method achieves performance on par with backpropagation through time (BPTT)~\cite{werbos1990backpropagation} on long-sequence modeling tasks. It inherits the same scalability limitations of DeepZero, as the need for individual parameter perturbations imposes heavy computational and infrastructural demands. Yet, RNNs are known for challenges like vanishing/exploding gradients and long-range dependencies, which makes ZO alternatives like this particularly noteworthy.

Recently, methods within the category of \textit{population-based metaheuristics}, in particular ES, have also been successfully used to train NNs in specific domains, exploiting the same principle of gradient-approximation. As mentioned earlier in this review, ES update has been shown to converge to finite-difference approximations of the gradient, indicating that ES approximates gradient descent~\cite{zhang_relationship_2017, lehman_es_2018, raisbeck_evolution_2019}. Specifically, ES have been used to train NNs from scratch in reinforcement learning (RL) environments~\cite{salimans2017evolution, such2017deep}. 

Despite the limitations of ZO methods in high-dimensional optimization, the most successful approaches for training large NNs share a common strategy. They combine stochastic perturbations to explore the parameter space with a global feedback signal to guide the optimization. This elegant paradigm approximates the principles of gradient descent without needing explicit gradients, proving to be both effective and scalable. This fundamental insight will lead us to the last section, where we will explore how these same optimization principles may be at play in biological systems, offering a new perspective on how the brain learns and adapts.

\subsection{Learning dynamics in overparameterized neural networks}

Overparameterized NNs, which have more parameters than data points, often show unique and counterintuitive learning dynamics. A central part of this is the resulting high-dimensional and non-convex loss landscape, which governs how the parameters are updated during training. 

Optimization methods like SGD consistently find solutions that generalize well, challenging the conventional belief that non-convex landscapes are filled with suboptimal local minima, which would prevent the optimization process from finding the best solution. Now, it is widely believed that this success is due to the geometry of these landscapes, where better solutions correspond to flat minima~\cite{hochreiter1997flat, keskar2016large, dziugaite2017computing}. These flat minima are characterized by a broad, low-error region in the parameter space, which makes the model more robust to perturbations and leads to better generalization. In contrast, sharp minima tend to result in poor generalization. The landscape of the train loss function is a learned empirical approximation of the true, underlying landscape, which is represented by the test loss function. Thus, these two landscapes are more similar on average in smoother areas, resulting in a better approximation of the true loss function. Figure \ref{fig:Fig2}b shows a visual example of how this affects generalization.

Techniques such as small-batch training (and its noisy gradient estimates)~\cite{keskar2016large}, dropout~\cite{srivastava2014dropout}, skip connections~\cite{he2016deep}, and batch normalization~\cite{ioffe2015batch} contribute to a smoother loss landscape with more abundant flat minima. In addition, recent methods have been proposed to explicitly seek parameters within these low-loss neighborhoods, encouraging the discovery of flatter minima~\cite{kaddour2022flat, li2025seeking}.

A further insight is that in the overparameterized regime, the set of global minima is not a single point, but a high-dimensional connected manifold in the parameter space~\cite{draxler2018essentially, garipov2018loss}. This connectivity suggests that different good solutions are not isolated but rather linked, making it easier for optimization algorithms to find a path to solutions that generalize better.

Moving beyond the general properties of the loss landscape, a key finding in the overparameterized regime is the double descent phenomenon~\cite{belkin2019reconciling}. This phenomenon challenges the classical U-shaped curve of test error, which shows that as model complexity (e.g., number of parameters) increases, the test error first decreases and then increases again due to overfitting. As model complexity continues to increase past the point where the model can perfectly fit the training data (the interpolation threshold), the test error paradoxically begins to decrease again, as shown in Figure \ref{fig:Fig2}c.

In the underparameterized regime (fewer parameters than data points), the model is constrained and cannot achieve zero training error. Its learning is governed by the classic bias-variance trade-off, where increasing complexity reduces bias but raises variance, leading to the first descent and eventual increase in test error. In contrast, in the overparameterized regime, the model becomes unconstrained and can find multiple solutions that perfectly fit the training data. At this point, the key to good generalization lies in the implicit bias of SGD, which is its inherent preference to find simpler solutions~\cite{gunasekar2018implicit}. The learning process in this regime can be seen as two phases: first, the model learns to perfectly interpolate the training data, and then refines its internal data representations using the implicit bias of SGD to find a more generalizable solution~\cite{nakkiran2021deep}.

Another striking phenomenon in the overparameterized regime is grokking~\cite{power2022grokking}, in which a delayed generalization occurs by a rapid decrease in the test error, long after its training error has reached zero, as shown in Figure \ref{fig:Fig2}c. This transition is often attributed to the model's two-phase learning process: an initial, rapid phase of memorizing the data, followed by a slower phase of generalization where the model discovers simpler solutions~\cite{liu2022towards}. During this second phase, the model refines its internal representations, shifting from a solution that simply memorizes the training samples to one that has learned a more efficient and robust rule for the given task. Regularization techniques such as weight decay~\cite{krogh1991simple, loshchilov2017decoupled}, dropout, and batch normalization facilitate grokking by slowly forcing the model to transition from a high-norm, memorizing solution to a simpler generalizing solution, thus controlling the timing and emergence of this delayed generalization. Interestingly, these regularization techniques can also mitigate or even completely eliminate the double descent effect due to the same inner workings on the learning dynamics.

These phenomena, from the geometry of the loss landscape to double descent and grokking, show that the learning dynamics of NNs are far more complex than previously thought, proving that our understanding of deep learning is still evolving. This is an active area of research with many exciting discoveries yet to be made.

\section{Biological learning: An optimization perspective}
\label{bio-learning}

Optimization provides a powerful framework for understanding adaptive systems, from engineered artificial intelligence (AI) algorithms to the complex emergent processes of learning in the brain. In this section, we extend the optimization principles and unique NNs challenges discussed earlier to reveal how similar computational concepts underlie biological learning mechanisms.

In artificial NNs, BP is the core algorithm for efficient gradient computation, driving deep learning success across a wide range of domains~\cite{jumper2021highly, gulshan2016development, mnih2015human, krizhevsky2012imagenet, degrave2022magnetic, merchant2023scaling, avsec2021effective}. BP's effectiveness has not only transformed AI but also inspired investigations into whether similar principles might operate in the brain. Interestingly, internal representations learned by BP-trained networks often resemble those found in neural recordings~\cite{cadieu2014deep, schrimpf2018brain, yamins2014performance, gucclu2015deep}, suggesting a possible convergence between artificial and biological learning systems. Despite these similarities, directly implementing BP in biological circuits presents serious challenges. Key issues include its reliance on non-local information, the weight transport problem (requiring symmetric forward and backward weights), the separation of inference and learning phases, and the need for neuron- or synapse-specific error signals in a non-differentiable biological substrate~\cite{whittington2019theories, lillicrap2020backpropagation, bengio2015towards}. These limitations have inspired research into biologically plausible alternatives for error propagation, giving rise to theories such as target propagation~\cite{bengio2014auto, lee2015difference, ahmad2020gait}, feedback alignment \cite{lillicrap2016random, nokland2016direct}, synthetic gradients~\cite{jaderberg2017decoupled}, local errors~\cite{nokland2019training}, and direct random target projection \cite{frenkel2021learning}.

Other alternative learning methods don't encode and propagate errors explicitly, but as activity differences over time or between different network states. These approaches include contrastive learning~\cite{o1996biologically, chen2020simple, oord2018representation}, equilibrium propagation~\cite{scellier2017equilibrium} (which bridges gradient-based learning with energy-based models like Hopfield networks~\cite{hopfield1982neural, ramsauer2020hopfield} and Boltzmann machines~\cite{ackley1985learning}), and the forward-forward algorithm~\cite{hinton2022forward}. In the domain of spiking neural networks (SNNs), alternative learning methods have also been proposed. These include SpikeProp~\cite{bohte2002error}, dendritic error signaling~\cite{sacramento2018dendritic}, dendritic event-based processing~\cite{yang2021efficient}, surrogate gradient learning~\cite{neftci2019surrogate}, e-prop~\cite{bellec2020solution}, burst synaptic plasticity~\cite{payeur2021burst}, and predictive coding~\cite{rao1999predictive}. Additionally, methods for converting artificial neural networks to SNNs have been developed to leverage the strengths of both paradigms~\cite{tavanaei2019deep}. While these methods are more experimental and ultimately aim at learning in artificial NNs, the next subsection will discuss neurophysiologically realistic learning theories of synaptic plasticity that are more widely accepted in the neuroscience community.

However, rather than replicating BP (or other specific-feedback-error method) exactly, biological learning may instead rely on more fundamental and widely applicable optimization principles. In particular, the combination of intrinsic stochasticity in neural systems and global reinforcement-like signals observed in the brain appears to support a form of learning that aligns remarkably with principles of ZO optimization. Our goal here is to connect this idea to the broader optimization landscape presented in this review.

The following subsections develop this perspective in detail. We begin by introducing the foundations of biological learning, providing essential background for readers without neurobiology experience. We then explore how learning could emerge through probing and reinforcing, supported by computational models and biological evidence. Finally, we integrate these insights to argue for an optimization view on biological learning, highlighting its implications for understanding natural and artificial intelligence.

\subsection{Foundations of biological learning}

Unlike the explicit algorithm-driven optimization in artificial systems, biological learning emerges from complex interactions of intricate biological circuits operating at different scales. Here, we outline these foundational aspects, as well as the emerging view of noise as a critical computational resource instead of a biological flaw. Rather than an exhaustive account, we briefly introduce these concepts to support further connections to optimization principles, directing readers to foundational texts for deeper study.

\subsubsection{Neurophysiological background}

\textbf{Synaptic plasticity}, the ability of synapses to strengthen or weaken over time, lies at the core of biological learning~\cite{citri2008synaptic}. It encompasses both long-term potentiation (LTP) and long-term depression (LTD)~\cite{malenka2004ltp}, widely considered the cellular mechanisms underlying lasting learning and memory. A core principle of plasticity is locality, the idea that synaptic changes are driven primarily by signals available at the synapse itself, such as local pre- and post-synaptic activity. This local learning is further shaped by neuromodulators like dopamine, serotonin, and acetylcholine~\cite{robbins1996neurobehavioural, kusmierz2017learning}, which diffuse across brain regions and act as global signals that modulate the expression of synaptic changes.

\textbf{Biological complexity}, however, goes far beyond these basic principles. For instance, neurons communicate via discrete spikes, enabling highly precise temporal processing~\cite{gerstner2014neuronal}; and short-term plasticity adjusts synaptic strength over milliseconds to seconds, supporting rapid information flow~\cite{zucker2002short}. Neural circuits also maintain a strict separation between excitatory and inhibitory synapses, whose balance is crucial for stable dynamics and computation~\cite{deco2014local}. Neurons are further organized into microcircuits for localized processing~\cite{douglas2004neuronal}, and learning capacity evolves over the lifespan, with critical periods of increased plasticity~\cite{hensch2004critical}. To connect these biological mechanisms with general optimization principles, we adopt a typical abstract, functional perspective that simplifies biological detail while preserving core concepts like locality and global modulation. For a deeper neurobiological background, see classic texts like~\cite{kandel2000principles, shepherd2004synaptic}. In the same vein, mechanisms of plasticity are diverse, but a few core mechanisms provide a strong foundation for understanding synaptic change at this level of abstraction. Hebbian learning~\cite{hebb2005organization} strengthens synapses when pre-synaptic activity drives post-synaptic firing. STDP~\cite{markram1997regulation, bi2001synaptic, legenstein2008learning} refines this by incorporating the precise timing of spikes. Calcium-based plasticity~\cite{malenka1988postsynaptic, graupner2012calcium} uses intracellular calcium levels to guide synaptic changes, while homeostatic plasticity~\cite{turrigiano2004homeostatic} ensures network stability by regulating excitability. Although other forms exist (e.g., inhibitory~\cite{castillo2011long}, structural~\cite{holtmaat2009experience}, and non-synaptic plasticity~\cite{daoudal2003long}), these are the most foundational from a computational perspective.

\begin{figure}[ht]
    \centering
    \includegraphics[width=\linewidth]{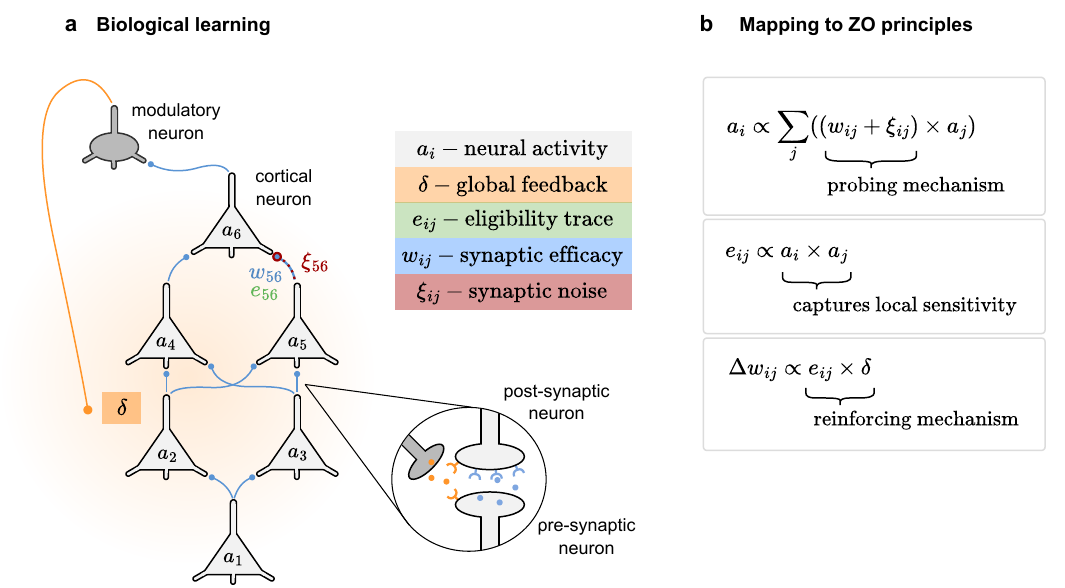}
    \caption{\textbf{Optimization and learning in biology.} \textbf{(a)} Overview of the core high-level elements involved in biological learning. Cortical neurons interact via synapses $w_{ij}$, whose efficacy is instantaneously perturbed by synaptic noise $\xi_{ij}$ --this noise can also be conceived as spontaneous neural firing. A global feedback signal $\delta$ is then delivered by a modulatory neuron (e.g., dopaminergic), which signals changes in performance or expectations (e.g., reward prediction error). Synaptic changes are mediated by eligibility traces $e_{ij}$, a local memory of past activity correlations. \textbf{(b)} Computational mapping to zeroth-order optimization principles. The first formula illustrates how the probing mechanism is integrated into the processing pathways. The second formula shows how each synapse is tagged with an eligibility trace, capturing local sensitivity and activity correlations. The final formula provides the reinforcing mechanism, estimating a descent direction based on changes in performance or expectation.    
    }
    \label{fig:Fig3}
\end{figure}

\textbf{Three-factor learning} provides a powerful framework that integrates these core principles. It extends the classic two-factor Hebbian principle by adding a third global modulatory signal that determines whether local synaptic changes are expressed~\cite{fremaux2016neuromodulated, kusmierz2017learning}. These three factors link fast local neuronal and synaptic changes to slower global outcomes, bridging the gap between neuronal and behavioral timescales. A common implementation involves eligibility traces~\cite{gerstner2018eligibility}, which flag synapses for potential change based on recent local activity. These traces are later modulated by delayed global signals, such as reward prediction errors from dopaminergic neurons~\cite{schultz1997neural, schultz2016dopamine}, which encode the mismatch between expected and received rewards. By combining local activity with delayed global feedback, three-factor learning overcomes the limitations of purely local rules and supports learning across timescales. Figure \ref{fig:Fig3}a provides a visual scheme of the core high-level elements involved in biological learning.

\subsubsection{Noise as a computational resource}

A paradigm shift in neuroscience is redefining neural noise, not as a biological flaw, but as a fundamental computational resource~\cite{faisal2008noise, maass2014noise, rolls2010noisy}. Although noise can disrupt signal transmission, increasing evidence shows that the brain may be leveraging stochasticity, from random neurotransmitter release to spontaneous neuron firing, for essential functions. This inherent variability introduces useful dynamics into neural networks, allowing them to represent knowledge as probability distributions and perform efficient probabilistic inference through sampling~\cite{Kappel2015Network}. Beyond this, neural noise also facilitates exploration by probing the state space, supporting the network's self-organization and enabling the system to escape local optima~\cite{faisal2008noise, maass2014noise}. In the context of decision-making, stochasticity is crucial for robust performance, preventing attractor networks from getting stuck in deterministic states and improving probabilistic choice behavior~\cite{rolls2010noisy}. Embracing this perspective helps to understand how the brain tackles complex problems and adapts to uncertainty in the real world.

Neuromorphic computing can, in turn, benefit enormously from this perspective. Neuromorphic devices are designed to mimic the brain's energy efficiency~\cite{kaspar2021rise, roy2019towards} by processing information in a massively parallel and decentralized manner. However, they are often inherently noisy due to their low-power operation and physical implementation (e.g.,~\cite{gaba2013stochastic, milano2022materia}). Viewing noise as a computational feature opens the door to learning methods that harness this inherent stochasticity for improved robustness and adaptability~\cite{koenders2025noise}.

\subsection{Computational models of learning by probing and reinforcing}

These elements of biological learning can then be framed computationally and linked to broader principles of optimization. At the core of this process is the interplay between local perturbations and global feedback: The brain's inherent noise naturally acts as a probing mechanism to explore state and parameter spaces, with learning enabled by a global reinforcement signal that guides adaptation. The network thus learns to associate local neural and synaptic changes with behavioral outcomes, reinforcing beneficial patterns and suppressing unproductive ones. Although seemingly random, this process parallels ZO optimization, where random explorations of the parameter space and performance feedback guide directions of improvement without explicit gradients. This computational mapping is schematically summarized in Figure \ref{fig:Fig3}b.

Recent computational models have instantiated these core principles at varying levels of biological detail to explain experimental data. Early works in this area, like~\cite{mazzoni_more_1991}, which introduced stochastic perturbations in model neurons, and~\cite{seung_learning_2003}, which examined probabilistic neurotransmitter release, paved the way for noise-driven learning frameworks. Subsequent work made the link to ZO methods more explicit. Fiete et al.~\cite{fiete_gradient_2006} showed that random fluctuations in membrane conductance can approximate gradient descent, later applying this insight to build a model of birdsong learning~\cite{fiete_model_2007}. Legenstein et al.~\cite{legenstein_reward-modulated_2010} proposed a reward-modulated Hebbian rule that maps to the NP algorithm and used it to explain experimental observations in a brain-computer interface task, which was later extended to incorporate more realistic delayed rewards~\cite{miconi_biologically_2017}. Anatomically detailed models, in the basal ganglia~\cite{bouvier_cerebellar_2018} and cerebellum~\cite{schmid_forward_2019}, further demonstrated how stochastic firing mechanisms can be used to implement stochastic gradient descent. On the other end of the spectrum, functionality-focused approaches, like Ma et al.~\cite{ma_exploiting_2023}, strictly derive a noise-driven rule that resembles three-factor Hebbian learning, validating it on complex deep network architectures and datasets, while Garcia et al.~\cite{garcia_fernandez_ornsteinuhlenbeck_2024} exploit continuous-time neurotransmitter stochasticity to learn both synaptic weights and optimal noise levels based on input and performance.

Collectively, these studies demonstrate that this approach to learning is a powerful and plausible mechanism for how the brain learns. This perspective is gaining increasing traction in the neuroscience community, with researchers arguing that the search for adaptive physiological changes in the brain is, in essence, a search for mechanisms that allow the brain to approximate gradients, a unifying idea that explains many facets of neuronal plasticity~\cite{richards2023study}.

\section{Conclusion and future directions}

This review paper provides a unified perspective on iterative optimization, bridging the gap between general theory, AI, and biological learning. We began by exploring the landscape of gradient-based optimization, from the foundational first-order methods to the more complex second- and higher-order approaches that leverage curvature information. Building on this, we then shifted our focus to derivative-free, or zeroth-order, methods, which operate without explicit gradients. We categorized these approaches, highlighting approximate-gradient methods and population-based metaheuristics that can effectively mimic gradient-based principles by employing random perturbations and feedback. This comprehensive overview provided a crucial bridge, allowing us to examine how these computational concepts apply to the unique, high-dimensional challenges of training neural networks. We emphasized that while backpropagation is the dominant force in AI, effective learning can also emerge from methods that simply approximate gradients. This core insight leads us to the final part of our discussion, where we argue that the very same principles likely underpin learning in the brain.

The increasing performance of non-gradient-based methods suggests a broadening of the optimization landscape in machine learning~\cite{liu_primer_2020}. Although backpropagation has long been the standard, recent breakthroughs show that scalable alternatives are emerging. For example, methods such as DeepZero~\cite{chen_deepzero_2023} and the work of Chaubard et al.~\cite{chaubard2025scaling} have demonstrated that zeroth-order approaches can achieve a performance comparable to backpropagation, even for models with billions of parameters. Similarly, evolutionary strategies have proven to be a highly effective and scalable alternative in complex reinforcement learning environments~\cite{salimans2017evolution, such2017deep}. This progress encourages a move beyond the singular focus on backpropagation and highlights that effective learning can be achieved through different paradigms.

Furthermore, these advancements reinforce the idea that stochasticity is a crucial computational resource \cite{maass2014noise, zur2009noise}. The noisy updates from mini-batching in SGD, as well as other regularization techniques, are known to help models converge to flat minima, leading to better generalization \cite{kaddour2022flat, li2025seeking, xie2020diffusion}. This phenomenon has also been observed in the learning dynamics of overparameterized networks, such as double descent~\cite{belkin2019reconciling} and grokking~\cite{power2022grokking}. By embracing this perspective, we can inject additional stochasticity into the optimization process (as is inherent in many zeroth-order methods) to discover more robust and generalizable solutions. This perspective is especially promising for neuromorphic computing~\cite{kaspar2021rise, roy2019towards}. These brain-inspired devices are highly parallel, decentralized, and inherently noisy; thus, forcing them into the synchronous, memory-heavy workflow of backpropagation is far from optimal~\cite{esser2015backpropagation}. A zeroth-order learning framework based on local stochastic probes and global reinforcement is a much more natural fit. By leveraging their intrinsic noise and massive parallelism, we can unlock the potential of neuromorphic hardware for fast and energy-efficient AI~\cite{koenders2025noise,VanGerven2025Neuromorphic}.

Viewing learning through an optimization lens can also offer a powerful framework for understanding biological systems: the brain may be employing an efficient form of zeroth-order optimization. This perspective offers a plausible theory by mapping its core optimization principles to known neural mechanisms. In this view, the inherent neural stochasticity~\cite{faisal2008noise, rolls2010noisy}, from probabilistic neurotransmitter release to spontaneous spiking, serves as the probing mechanism that explores the high-dimensional space of synaptic weights~\cite{maass2014noise}. This natural variability allows the system to explore new solutions and escape local optima without precise deterministic gradients. Global neuromodulatory signals, such as dopamine-coded reward prediction errors~\cite{schultz1997neural, schultz2016dopamine}, then provide the reinforcement that guides local synaptic changes~\cite{fremaux2016neuromodulated, kusmierz2017learning}. This scheme, which combines local stochastic exploration with a global performance feedback signal, parallels the successful zeroth-order methods reviewed.

This view has the potential to integrate disparate observations: adaptive physiological changes may represent mechanisms for approximating gradients \cite{richards2023study}. In essence, it suggests that the brain's solution to the learning problem is a robust, decentralized, and noise-driven process that finds good solutions by probing and reinforcing. Ultimately, this view encourages neuroscientists to shift their focus from the search for the brain's learning algorithm to an investigation of the fundamental optimization principles that guide its adaptive behavior~\cite{richards2019deep, jonas2017could}.

\bibliographystyle{abbrv}
\bibliography{references}

\end{document}